\RequirePackage{fix-cm}

\documentclass{svjour3}                 

\smartqed

\usepackage[T1]{fontenc}
\usepackage[utf8]{inputenc} 
\usepackage{times}          

\usepackage{graphicx}
\usepackage{float}
\usepackage{booktabs}
\usepackage{multirow}
\usepackage{multicol}
\usepackage{makecell}
\usepackage{tabularx}
\usepackage{pdflscape}
\usepackage[caption=false]{subfig}
\usepackage{rotating}
\graphicspath{{images/}{code_for_figures/images/}}
\usepackage{pdflscape} 
\usepackage{hyphenat}
\usepackage{tabularx}
\usepackage{placeins}
\newcolumntype{Y}{>{\centering\arraybackslash}X}
\newcolumntype{P}[1]{>{\raggedright\arraybackslash}p{#1}}
\newcolumntype{C}[1]{>{\centering\arraybackslash}m{#1}}
\newcolumntype{L}{>{\raggedright\arraybackslash}X}

\usepackage{amsmath}
\usepackage{amssymb}
\usepackage{mathtools}
\usepackage{icomma}

\usepackage[hidelinks]{hyperref}

\makeatletter
\providecommand\cl@chapter{}
\let\cl@chapter\@empty      
\makeatother

\usepackage[capitalize,nameinlink,noabbrev]{cleveref}

\crefname{section}{Section}{Sections}
\Crefname{section}{Section}{Sections}
\crefname{equation}{Eq.}{Eqs.}
\Crefname{equation}{Equation}{Equations}
\crefname{figure}{Fig.}{Figs.}
\Crefname{figure}{Figure}{Figures}
\crefname{table}{Tab.}{Tabs.}
\Crefname{table}{Table}{Tables}

\usepackage{algorithmic}
\usepackage{algorithm}

\usepackage{xcolor}
\definecolor{color_contrib}{HTML}{A2142F}
\definecolor{color_cond1}{HTML}{FFAF00}
\definecolor{color_cond2}{HTML}{00B3C4}

\usepackage{tikz}
\usetikzlibrary{shapes.geometric, arrows, positioning}

\usepackage{mdframed}
\usepackage{tcolorbox}
\tcbuselibrary{breakable, skins}
\definecolor{boxcolor}{rgb}{0.52, 0.18, 0.13}

\newtcolorbox[auto counter, number within=section]{infobox}[1]{
    colback=white,
    coltitle=black,
    boxrule=0pt,
    borderline west={3pt}{0pt}{boxcolor},
    enhanced jigsaw,
    left=10pt,
    right=0pt, top=0pt, bottom=0pt,
    fonttitle=\bfseries,
    title={#1},
    colframe=white,
    before=\par\medskip,
    after=\par\medskip,
    breakable,
    parbox=false
}

\usepackage{url}
\usepackage{xurl}

\usepackage[savewrites,seeautonumberlist]{glossaries}
\makenoidxglossaries

\newacronym{HDVSA}{HDC/VSA}{Hyperdimensional Computing a.k.a. Vector Symbolic Architectures}
\newacronym{HDC}{HDC}{Hyperdimensional Computing}
\newacronym{VSAs}{VSAs}{Vector Symbolic Architectures}
\newacronym{HD}{HD}{high-dimensional}
\newacronym{FHRR}{FHRR}{Frequency Holographic Reduced Representation}
\newacronym{ROCKET}{ROCKET}{RandOm Convolutional KErnel Transform}
\newacronym{PPV}{PPV}{Proportion of Positive Values}
\newacronym{MPV}{MPV}{Mean of Positive Values}
\newacronym{MIPV}{MIPV}{Mean Index of Positive Values}
\newacronym{LSPV}{LSPV}{Longest Stretch of Positive Values}
\newacronym{LOO-NMSE}{LOO-NMSE}{Leave-One-Out Normalized Mean Squared Error}
\newacronym{FPE}{FPE}{Fractional Power Encoding}
\newacronym{ACC}{ACC}{Accuracy}
\newacronym{AUCROC}{AUC-ROC}{Area Under the Receiver Operating Characteristic Curve}
\newacronym{LOOCVNMSE}{LOOCV-NMSE}{Leave-One-Out Negative Mean Squared Error}
\newacronym{TSC}{TSC}{Time Series Classification}
\newacronym{ANN}{ANN}{Artificial Neural Network}
\newacronym{ANNs}{ANNs}{Artificial Neural Networks}
\newacronym{RNN}{RNN}{Recurrent Neural Network}
\newacronym{RNNs}{RNNs}{Recurrent Neural Networks}
\newacronym{SNN}{SNN}{Spiking Neural Network}
\newacronym{SNNs}{SNNs}{Spiking Neural Networks}
\newacronym{LSTM}{LSTM}{Long Short-Term Memory}
\newacronym{LIF}{LIF}{Leaky-Integrate-and-Fire}
\newacronym{GRU}{GRU}{Gated Recurrent Unit}
\newacronym[plural=VSAs, longplural=Vector Symbolic Architectures]{VSA}{VSA}{Vector Symbolic Architecture}
\newacronym{RFF}{RFF}{Random Fourier Features}

\newacronym{SPA}{SPA}{Semantic Pointer Architecture}
\newacronym{SVM}{SVM}{Support Vector Machine}
\newacronym{kNN}{kNN}{k-nearest neighbors}
\newacronym{ADAS}{ADAS}{Advanced Driver Assistance Systems}

\newacronym{BSDC}{BSDC}{Binary Sparse Distributed Code}
\newacronym{HRR}{HRR}{Holographic Reduced Representation}
\newacronym{MAP}{MAP}{Multipy-Add-Permute}
\newacronym{BSC}{BSC}{Binary Spatter Code}

\newglossaryentry{predictor}{
    name=predictor,
    description={is a set of attributes or features of the data that are used to predict the class label or category of the data instances and can be represented as a vector (feature vector)}
}
\newglossaryentry{encoder}{
    name=encoder,
    description={is a function that maps the input data to a feature vector. In the context of time series classification, the encoder is used to transform the input time series into a feature vector that can be used for classification}
}
\newglossaryentry{aggregation}{
    name=aggregation,
    description={is the process of combining multiple data points into a single representation (e.g. based on summing, averaging, etc.). In the context of time series classification, for example, aggregation is used to create a representation from the feature vectors over time by summing}
}
\newglossaryentry{filter-kernel}{
    name=filter kernel,
    description={is a set of weights that are used to convolve the input data to create features}
}
\newglossaryentry{kernel}{
    name=kernel,
    description={is a function that measures the similarity between two objects (vectors) from a set of objects}
}
\newglossaryentry{global}{
    name=global,
    description={is a property that refers to the entire time series data}
}
\newglossaryentry{local}{
    name=local,
    description={is a property that refers to a subset of the time series data and is used to capture relationships in neighboring data points as in convoltions}
}

\makeatletter
\def\makeheadbox{}
\makeatother

\begin{document}

\title{
Real-Valued Hyperdimensional Sequence Representations with Hadamard Product Binding and Shift Equivariance
}
\titlerunning{
Real-Valued Sequence Representations with Hadamard Binding
}

\author{
Kenny Schlegel \and
Dmitri A. Rachkovskij \and
Denis Kleyko \and
Amy Loutfi \and
Stefan Streif \and
Evgeny Osipov
}

\institute{
K. Schlegel \at Chemnitz University of Technology, Germany \\
\email{kenny.schlegel@etit.tu-chemnitz.de}
\and
D. A. Rachkovskij \at Luleå University of Technology, Sweden \\
Institute of Information Technologies and Systems, Ukraine
\and
D. Kleyko \at Örebro University, Sweden \\
RISE Research Institutes of Sweden, Sweden
\and
A. Loutfi \at Örebro University, Sweden \\
Linköping University, Sweden
\and
S. Streif \at Chemnitz University of Technology, Germany
\and
E. Osipov \at Luleå University of Technology, Sweden
}

\date{}

\maketitle

\begin{abstract}
Encoding temporal order is a fundamental requirement for sequence representations in Hyperdimensional Computing.
Fractional Power Encoding provides similarity-preserving position vectors whose inner products approximate shift-invariant kernels, and it supports shift-equivariant transformations of encoded sequence representations. However, standard formulations of Fractional Power Encoding are primarily designed for binding operations such as circular convolution or complex-valued multiplication, which limits their compatibility with Hadamard product binding of real-valued vectors.
This paper develops real-valued position encodings motivated by Random Fourier Features, aiming to retain the desirable properties of Fractional Power Encoding while supporting Hadamard-based operations. We propose three real-valued position-encoding variants: a real-valued baseline based on the inverse Fourier transform, and Sinusoid and Cosine-only representations derived from Random Fourier Features.  
Among them, the Sinusoid variant provides an explicit algebraic shift operator, allowing temporal shifts to be applied directly to the vector-encoded sequence representation without re-encoding the shifted sequence.
Experiments on time-series classification datasets show that the proposed real-valued representations achieve performance comparable to standard Fractional Power Encoding while enabling computationally efficient Hadamard product binding. 
The Sinusoid variant offers the most favorable trade-off, combining efficient real-valued implementation with exact shift-equivariant transformations.
\end{abstract}


\newcommand{\had}{\odot}
\newcommand{\shiftop}{\diamond}

\section{Introduction}
\label{sec:intro}

Encoding the order of elements (features or entities) in sequences is an important capability of \gls{HDVSA}. Sequence elements can be combined into a compact, fixed-size vector representation that preserves temporal structure through binding and superposition. 
Throughout this work, the term vector refers to high-dimensional vectors, typically with hundreds or thousands of components. 
In a typical encoding approach, each element's vector is bound with its position in the sequence, and the resulting bound vectors are superimposed to form a single high-dimensional representation of the entire sequence. A central question is how element--position bindings are constructed. Many approaches encode position ``implicitly'' by applying position-dependent permutations to element vectors~\cite{Kussul1991,Kanerva2009,Rachkovskij2024}.
In this paper, we instead consider approaches in which positions are represented explicitly by position vectors~\cite{Rachkovskij1990,Plate1994}. 
Such position vectors allow the sequence representation to capture similarity or proximity relationships between nearby positions. 

One approach to constructing such position vectors is \gls{FPE} \cite{Frady2021a,Plate1994,Komer2019,Komer2020,Lu2019}. 
FPE generates position vectors whose similarity reflects the distance between positions, thereby inducing graded similarity kernels over time. It also supports shift\hyp equivariant sequence representations: a temporally shifted sequence can be represented by transforming the already encoded sequence representation, rather than by recomputing all element--position bindings at the shifted positions~\cite{Rachkovskij2022a,Rachkovskij2022b,Rachkovskij2024}.
This property is useful because temporal patterns often occur at different positions within a sequence.
It is relevant for comparing temporally shifted patterns, searching over possible alignments, and updating sequence representations in streaming settings where newly arriving elements shift previous elements into the past. 
Existing implementations of \gls{FPE} rely on binding operations used in \gls{HRR} and \gls{FHRR} \gls{HDVSA} models. 
These models use real-valued circular convolution or component-wise multiplication in complex-valued space as binding operations~\cite{Plate1994,Frady2021}. Although this formulation provides the desired shift equivariance, efficient implementation of circular convolution typically relies on Fourier transforms, which introduces additional computational overhead.

At the same time, efficient real-valued \glspl{HDVSA} models, such as \gls{MAP}, use the Hadamard product (i.e., component-wise multiplication) as the binding operation \cite{Gayler1998a}. 
However, it has been noted that \gls{FPE} cannot be directly implemented with the real-valued Hadamard product because the required algebraic structure is not available in real-valued vector spaces~\cite{Frady2021}. 
As a result, existing \gls{FPE}-based approaches are not directly compatible with efficient real-valued \glspl{HDVSA} models.

This paper addresses this limitation by investigating real-valued position encodings that are compatible with Hadamard product binding while retaining the desirable properties associated with \gls{FPE}, including similarity structure and shift equivariance. 
The following sections first review the relevant background and then introduce several real-valued position-encoding variants, together with their theoretical properties and empirical evaluation.

\section{Related Work}
\label{sec:related}
A central goal in \gls{HDVSA} sequence representation  is to encode a sequence of vectors 
$\mathbf{s} = [\mathbf{f}_1, \mathbf{f}_2, \dots, \mathbf{f}_T]$ 
into a single fixed-size vector representation. 
Two common approaches are position-dependent permutation and binding with explicit position vectors.

In \emph{permutation-based encoding}, each sequence element is transformed by a position-dependent permutation $\rho^t$ and then superimposed:
\begin{equation}
    \mathbf{r} = \rho^1(\mathbf{f}_1) + \rho^2(\mathbf{f}_2) + \dots + \rho^T(\mathbf{f}_T).
\end{equation}
This approach captures order information. However, when random permutations are used, it does not preserve similarity between nearby positions: \(\rho^t(\mathbf{x})\) and \(\rho^{t+1}(\mathbf{x})\) are nearly orthogonal~\cite{Kanerva2009}. Several extensions address this limitation, including partial permutations~\cite{Kussul2006} and more recent shift-equivariant encodings that enforce gradual similarity changes across nearby positions~\cite{Rachkovskij2024}.

In explicit \emph{role--filler encoding}, each element vector is bound to an explicit position vector $\mathbf{p}_t$ before superposition:
\begin{equation}
    \mathbf{r} = \mathbf{p}_1 \circ \mathbf{f}_1 + \mathbf{p}_2 \circ \mathbf{f}_2 + \dots + \mathbf{p}_T \circ \mathbf{f}_T,
    \label{eq:sequence_enc_bind}
\end{equation}
where $\circ$ denotes the binding operation. 
This formulation is common in \gls{HDVSA} and allows the properties of the sequence representation to be controlled through the construction of the position vectors~\cite{Rachkovskij1990,Plate1994}. 
In particular, similarity-preserving position encodings make it possible to assign similar position vectors to nearby positions, thereby capturing proximity relationships within the sequence.

A widely used example of such encodings is Fractional Power Encoding (\gls{FPE}), originally introduced in the context of \gls{HRR}~\cite{Plate1994}. 
\gls{FPE} generates position vectors by exponentiating a base vector in the complex domain, producing a family of vectors whose similarity decreases gradually with the increased positional difference. 
This mechanism induces a shift-invariant similarity kernel over the encoded position variable. 

Related work has also described this approach as Kernel Locality Preserving Encoding (KLPE), emphasizing its interpretation as an explicit kernel feature map~\cite{Frady2021a}. 
In this view, the inner product between encoded positions approximates a shift-invariant similarity kernel whose shape is determined by the distribution of sampled angular frequencies.
This perspective connects \gls{FPE} to Random Fourier Features (RFF)~\cite{Rahimi2008}, which approximate shift-invariant kernels using sampled frequencies and trigonometric feature maps. 

Both approaches encode positions through random features whose inner products approximate kernel similarities. 
However, in \gls{HDVSA}, \gls{FPE} is used to construct position vectors for binding-based sequence representations, whereas RFF is typically introduced as a general kernel approximation technique.

Most existing \gls{FPE} formulations are designed for \gls{HRR} and \gls{FHRR}, where binding is implemented through circular convolution or complex-valued component-wise multiplication~\cite{Plate1994,Frady2021a}. 
In contrast, efficient real-valued \gls{HDVSA} models such as \gls{MAP} use the Hadamard product as the binding operation. 
Although \gls{FPE}-generated position vectors have already been used together with Hadamard product binding in practical applications, for example, in time-series classification with \gls{HDVSA}~\cite{Schlegel2022b,Schlegel2025a}, the compatibility between \gls{FPE}-based position vectors and real-valued Hadamard product binding, including the resulting similarity and shift-equivariance properties, has not been systematically analyzed.

The Sinusoid variant studied in this work is also conceptually related to Rotary Positional Encoding (RoPE) used in transformer neural networks~\cite{Su2024}. 
Both approaches encode positions through trigonometric phase representations and express relative shifts as transformations in the embedding space. 
However, RoPE integrates positional information into the attention mechanism, whereas the present work constructs explicit position vectors for binding-based sequence representations in \gls{HDVSA}.

In this article, we, therefore, focus on sequence encodings based on multiplicative binding, as in \cref{eq:sequence_enc_bind}, and investigate how similarity-preserving position encodings with shift-equivariant properties can be realized efficiently in real-valued vector spaces using the Hadamard product operations. 
Permutation-based approaches and related \(n\)-gram encodings~\cite{Kleyko2023a,Kleyko2023b} are not considered further.
\section{Preliminaries}
\label{sec:prelim}

\subsection{Binding-based sequence encoding}

We consider sequence representations based on multiplicative binding of element vectors with explicit position vectors. 
Given a sequence of element vectors $\mathbf{f}_t \in \mathbb{R}^D$ at positions $t=1,\dots,T$, the sequence representation is constructed as

\begin{equation}
\mathbf{r} = \sum_{t=1}^{T} \mathbf{p}_t \circ \mathbf{f}_t ,
\label{eq:seq_enc}
\end{equation}
where $\mathbf{p}_t$ denotes the position vector at time step $t$ and $\circ$ denotes the binding operation used by the underlying \gls{HDVSA}. 
This formulation corresponds to the role--filler encoding introduced in \cref{sec:related}. 
The properties of the resulting sequence representation depend largely on how the position vectors $\mathbf{p}_t$ are constructed.

\subsection{Fractional Power Encoding}
\label{subsec:orig_fpe}

Fractional Power Encoding (\gls{FPE}) provides a similarity-preserving approach for generating position vectors. Instead of assigning independent vectors to different positions, \gls{FPE} constructs a family of related position vectors from a single base vector through repeated binding operations:

\begin{equation}
\mathbf{p}_t 
= \underbrace{\mathbf{p} \,\circledast\, \mathbf{p} \,\circledast\, \cdots \,\circledast\, \mathbf{p}}_{t\ \text{times}}
= \mathbf{p}^{(\circledast t)}
= \mathrm{IFFT}\!\left( (\mathrm{FFT}(\mathbf{p}))^t \right).
\label{eq:conv_power}
\end{equation}

In the standard formulation of \gls{HRR}~\cite{Plate1994}, binding is realized by circular convolution $\circledast$, which corresponds to component-wise multiplication in the complex-valued Fourier domain using the $\mathrm{FFT}$ and $\mathrm{IFFT}$, as shown in \cref{eq:conv_power}.
Let $\mathbf{p}\in\mathbb{R}^D$ denote a base vector and let $\mathbf{c}=\mathrm{FFT}(\mathbf{p})$ be its Fourier transform. 

The base vector can be constructed in different ways. 
In the original \gls{HRR} formulation, $\mathbf{p}$ is sampled from a Gaussian distribution and normalized to unit magnitude before applying the Fourier transform.
 Alternatively, the Fourier-domain representation can be constructed directly by sampling 
 angular frequencies \(\omega_j\) from a chosen distribution \(\mathcal{P}\) and defining unit-magnitude complex components \(e^{i\omega_j}\). At position \(t\), each Fourier component then evolves as
\begin{equation}
    c_j(t)
    =
    e^{i\omega_j t}.
\end{equation}
Thus, each Fourier component can be interpreted as a complex phasor rotating on the unit circle with angular frequency \(\omega_j\).

The phasor-based formulation makes the kernel properties of \gls{FPE} explicit, since the distribution \(\mathcal{P}\) of angular frequencies determines the resulting similarity kernel of the position vectors, as discussed in \cref{subsec:sim_kernel}. When real-valued position vectors are required, the Fourier coefficients must satisfy Hermitian symmetry,
\begin{equation}
c_k = \overline{c_{D-k}},
\end{equation}
so that the inverse Fourier transform yields a real-valued vector in the spatial domain \cite[p.~80]{Plate1994}. In contrast, when operating directly in the complex domain, as in \gls{FHRR}, this constraint is not necessary because no inverse Fourier transform is applied.

Assuming an even dimensionality \(D\) for simplicity, and using a scale parameter \(\beta\) to control the similarity decay between nearby positions, the Hermitian-symmetric Fourier-domain position vector can be written as

\begin{equation}
\mathbf{c}(t,\beta)
=
\left[
1,\,
e^{i\omega_2\beta t},\,
\dots,\,
e^{i\omega_{D/2}\beta t},\,
1,\,
e^{-i\omega_{D/2}\beta t},\,
\dots,\,
e^{-i\omega_2\beta t}
\right],
\label{eq:fpe_function}
\end{equation}
where the angular frequencies \(\omega_j\) of the sampled Fourier components are drawn i.i.d. from \(\mathcal{P}\), for example, \(\mathcal{U}(-\pi,\pi)\). The entries equal to \(1\) are fixed real-valued components required by the Hermitian-symmetric construction.

The corresponding real-valued position vectors used in \gls{HRR} are obtained through the inverse Fourier transform:
\begin{equation}
\mathbf{p}(t,\beta)
=
\mathrm{IFFT}(\mathbf{c}(t,\beta)).
\label{eq:fpe_real_ifft}
\end{equation}

This formulation extends repeated binding to continuous values of \(t\), enabling graded similarity-preserving position encodings. Hermitian symmetry ensures that \(\mathbf{p}(t,\beta)\) is real-valued, while the unit-magnitude complex components in \cref{eq:fpe_function} ensure constant norms of the position vectors for all \(t\).

\begin{figure}[tb]
    \includegraphics[width=\textwidth]{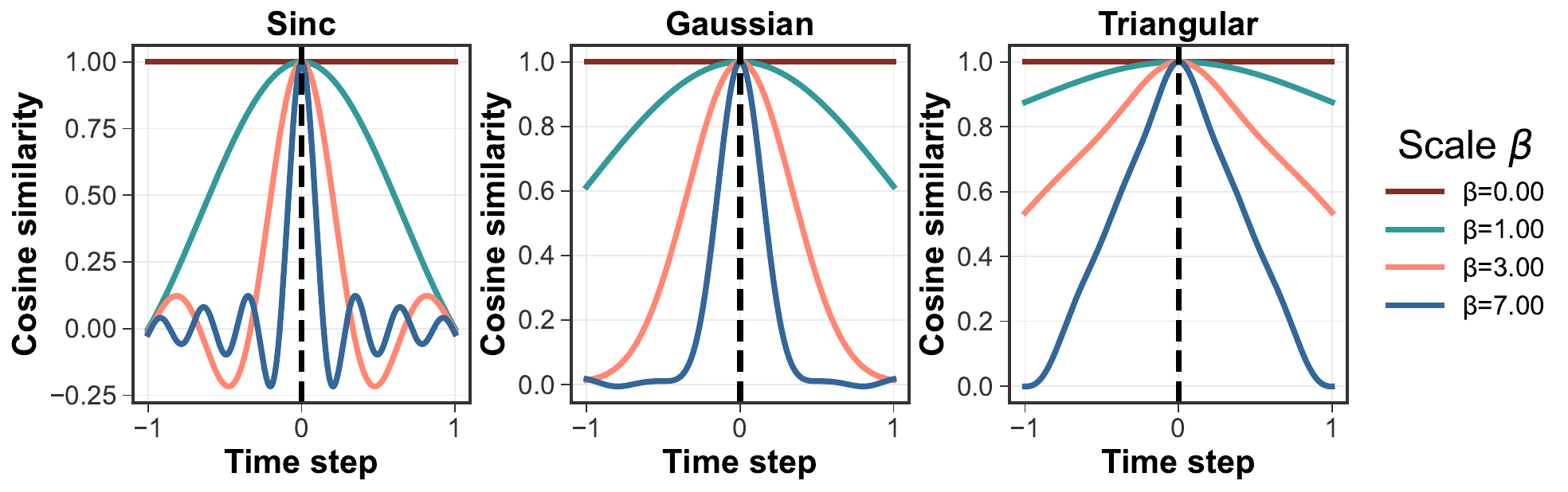}
    \caption[Similarity kernel for different FPE frequency distributions]{
    Similarity kernel \(\kappa(t,0)\) for different angular-frequency distributions used to generate \gls{FPE} position vectors and for different width parameters \(\beta\). 
    Each curve shows the similarity between the encoding at position \(t\) and the reference position \(t=0\). 
    Uniform sampling of angular frequencies produces a sinc-type kernel, Gaussian sampling yields a Gaussian-like kernel, sampling from triangular frequency distributions produces compact-support kernels.    
    }
    \label{fig:similarity_timesteps}
\end{figure}

\subsection{Similarity kernel}
\label{subsec:sim_kernel}

An important property of \gls{FPE} is that the similarity between position vectors depends only on the difference between their positions. 
More generally, a kernel function measures similarity between elements of a set. 
In the case of \gls{FPE}, the inner product between position vectors approximates
certain shift-invariant kernel functions $k(t_1,t_2) = k(t_1-t_2)$ over the time steps $t_1$ and $t_2$~\cite{Frady2021a,Kymn2023,Kymn2024}, meaning that the similarity depends only on the temporal distance between the two positions.
Specifically, since \(\|\mathbf{p}(t,\beta)\|^2=D\) for all \(t\), the similarity between two encoded positions is given by the normalized inner product:

\begin{equation}
\kappa(t_1,t_2)
=
\frac{1}{D}
\left\langle
\mathbf{p}(t_1,\beta),
\mathbf{p}(t_2,\beta)
\right\rangle .
\label{eq:kernel}
\end{equation}

This similarity approximates a shift-invariant kernel value,
\begin{equation}
\kappa(t_1,t_2) \approx k(t_1-t_2),
\label{eq:kernel_shift_equi}
\end{equation}
with the approximation improving as the dimensionality \(D\) increases. The shape of the approximated kernel \(k(t_1-t_2)\) is determined by the angular-frequency distribution \(\mathcal{P}\).

Examples of the resulting similarity kernels are shown in \cref{fig:similarity_timesteps}. Uniform sampling of angular frequencies produces a sinc-type kernel, Gaussian sampling yields a Gaussian-like kernel, and sinc-based frequency distributions produce kernels with linear (triangular) decrease.

\subsection{Shift equivariance}

Another important property of \gls{FPE} is shift equivariance. 
For a shift $\delta$, the following relation holds:

\begin{equation}
\mathbf{p}(t+\delta,\beta)
=
\mathbf{p}(\delta,\beta)
\circledast
\mathbf{p}(t,\beta),
\label{eq:fpe_equiv}
\end{equation}
where $\circledast$ denotes circular convolution. 
This relation follows directly from the exponentiation property in the complex domain,

\begin{equation}
\mathrm{FFT}(\mathbf{p}(t+\delta,\beta))
=
\mathbf{c}^{\beta(t+\delta)}
=
\mathbf{c}^{\beta t}
\cdot
\mathbf{c}^{\beta \delta}.
\end{equation}

\begin{figure}
  \centering
    \includegraphics[width=0.9\textwidth]{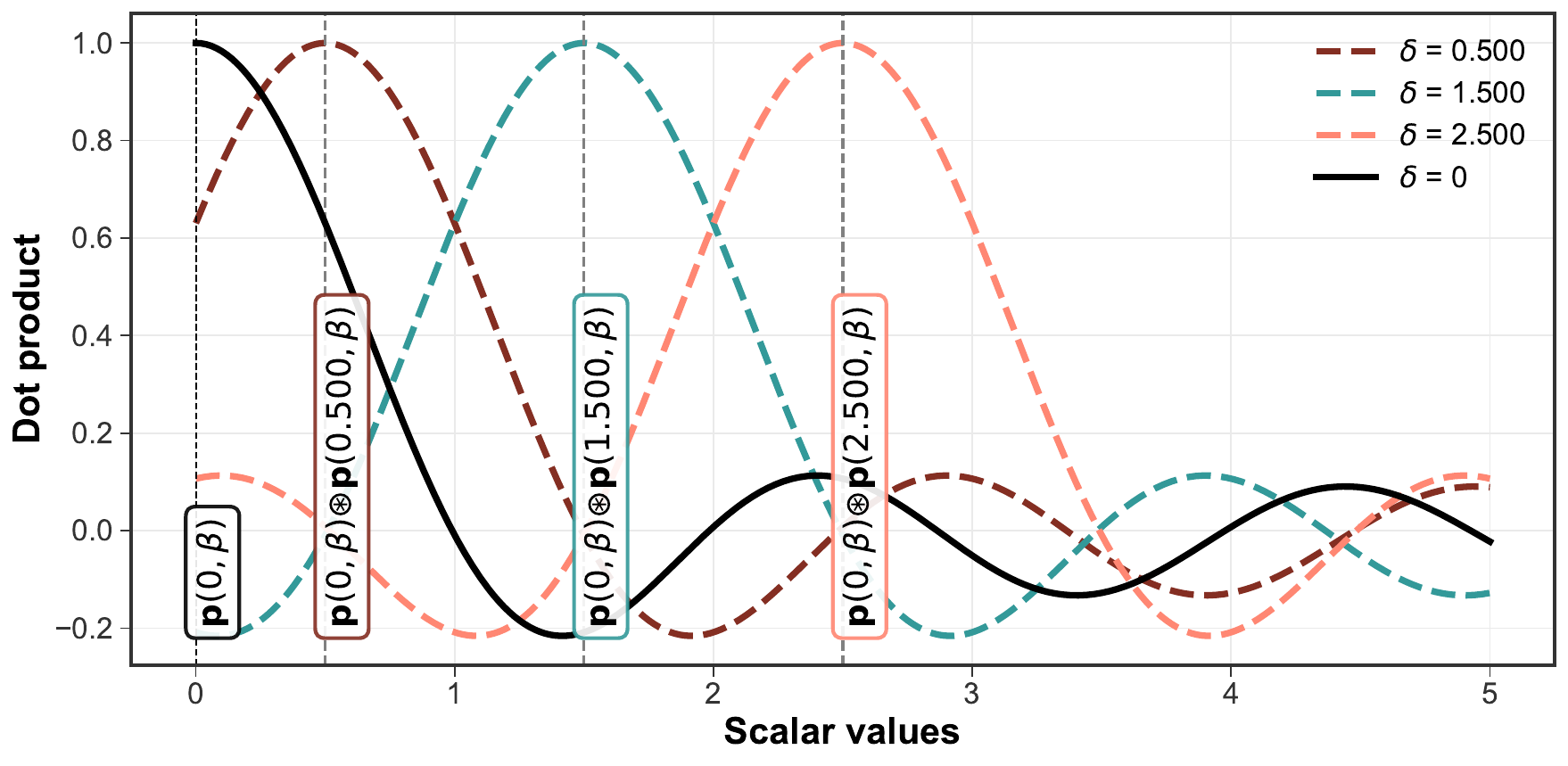}
    \caption[Shifted similarity kernels showing FPE equivariance]{
Shifted similarity kernels illustrating the equivariance of \gls{FPE}. 
Each curve shows the similarity between the shifted position vector
\(\mathbf{p}(\delta,\beta)\circledast\mathbf{p}(0,\beta)\)
and the position vectors \(\mathbf{p}(t,\beta)\) over different positions \(t\).
}
    
    \label{fig:fpe_shift_equivariance}
\end{figure}

Consequently, shifts in the encoded variable correspond to binding operations in the high-dimensional space. 
This property allows shifted sequence representations to be obtained directly from an existing representation without recomputing all element-position
bindings.
An example illustrating this behavior is shown in \cref{fig:fpe_shift_equivariance}. 
A shift in the position results in a corresponding shift of the similarity profile while preserving the kernel shape.

\section{Design Requirements}
\label{sec:req}

The previous section reviewed the standard formulation of \gls{FPE}, which provides similarity-preserving position vectors and shift-equivariant sequence representations. 
However, this formulation relies on circular convolution, or equivalently component-wise multiplication in the complex domain, as the binding operation. 
 These operations are not directly compatible with efficient real-valued \gls{HDVSA} models such as Multiply-Add-Permute (MAP)~\cite{Gayler1998a}, which use the Hadamard product as binding. This motivates the following design requirements for real-valued position encodings compatible with Hadamard product binding.

\paragraph{R1: Compatibility with Hadamard product:}

The position encoding must operate in real-valued vector space and support Hadamard product binding, $\mathbf{a} \odot \mathbf{b}
=
(a_1 b_1,\dots,a_D b_D) $, which is the standard binding operation in efficient real-valued \gls{HDVSA} models such as MAP.

\paragraph{R2: Similarity-preserving position vectors:}

The generated position vectors should retain the shift-invariant similarity structure associated with \gls{FPE}, meaning that the similarity between two position vectors depends on their positional difference, as shown in \cref{eq:kernel_shift_equi}. This property ensures that nearby positions produce similar vectors, whereas distant positions remain distinguishable.

\paragraph{R3: Stable vector norms:}

The norm of the position vectors should remain (exactly or approximately) constant with respect to $t$. 
Uncontrolled changes of position vector norms would lead to unstable sequence representations and distort the relative contribution of different sequence elements during superposition.

\paragraph{R4: Shift equivariance:}

Standard \gls{FPE} provides shift equivariance because a position shift is realized by binding with the vector representing the corresponding shift, as shown in \cref{eq:fpe_equiv}. However, in real-valued \gls{HDVSA} models such as MAP, where binding is implemented with the Hadamard product, binding with a ``shift vector'' does not automatically produce the encoding of the shifted position.

Therefore, a desirable property is the existence of an operator \(\diamond\) that performs the positional shift directly in the representation space:
\begin{equation}
\mathbf{p}(t+\delta)
=
\mathbf{p}(t)
\diamond
\mathbf{p}(\delta).
\end{equation}

This property allows temporal shifts to be applied directly to the vector representation of the whole sequence, without recomputing all element--position bindings at the shifted positions. Such behavior is useful for temporal alignment, shift search, streaming sequence processing, and efficient comparison of patterns occurring at different temporal positions.

\paragraph{R5: Computational efficiency:}

Finally, the encoding should avoid computationally expensive operations,
such as circular convolution with direct computational complexity $O(D^2)$, or $O(D \log D)$ complexity when implemented
using FFT and inverse FFT operations.
Ideally, position vectors should be computed directly in $\mathbb{R}^D$ with $O(D)$ complexity, and positional binding and shift operations should also be realized in $O(D)$ time.  \\

The following section introduces several real-valued position-encoding variants that satisfy these requirements to different degrees while enabling sequence encodings based on Hadamard product binding.

\section{Methods: Hadamard-Compatible Positional Representations}
\label{sec:methods}

This section introduces three real-valued positional representation variants compatible with Hadamard product binding. 
We encode a sequence of element vectors \(\{\mathbf{f}_t\}_{t=1}^{T}\) as
\begin{equation}
\mathbf{r} = \sum_{t=1}^{T} \mathbf{f}_t \odot \mathbf{p}(t,\beta),
\label{eq:method_seq_enc}
\end{equation}
where \(\mathbf{p}(t,\beta)\) is a position vector and \(\odot\) denotes component-wise multiplicative binding, i.e., Hadamard product binding.

The standard \gls{FPE} formulation reviewed in \cref{sec:prelim} defines position vectors by exponentiation in the complex domain and maps them to \(\mathbb{R}^D\) via the inverse Fourier transform. This construction yields a controlled shift-invariant similarity kernel, as in \cref{eq:kernel}, and supports shift equivariance under circular convolution, as in \cref{eq:fpe_equiv}, but requires Fourier-domain operations. 
Our goal is to obtain real-valued positional representations that retain the similarity-preserving structure associated with \gls{FPE}, support Hadamard product binding, and avoid Fourier transforms such as \cref{eq:fpe_real_ifft}. 
When possible, we also seek an explicit shift operator directly in the real-valued representation space.

The key observation is that the complex-domain formulation of \gls{FPE} can be interpreted through complex exponentials whose inner products approximate shift-invariant similarity kernels. 
This connection has previously been discussed in~\cite{Frady2021}, while Random Fourier Features~\cite{Rahimi2008} provide a related real-valued construction in which sampled angular frequencies are mapped to sine and cosine features. This perspective motivates the real-valued formulations studied here, which replace complex-domain representations with trigonometric maps and, in the Sinusoid variant, enable explicit real-valued shift operators based on phase-angle addition.

The variants differ in how \(\mathbf{p}(t,\beta)\) is constructed and whether an explicit shift operator exists. 
Variant~A follows the inverse-Fourier formulation as a real-valued baseline. 
Variant~B uses a Sinusoid RFF-based representation with an explicit shift operator. 
Variant~C uses a Cosine-only RFF-based representation, providing a simpler kernel approximation but without an algebraic shift operator.


\subsection[Real-valued standard FPE and Hadamard product]
{Variant A: Real-valued \gls{FPE} positional representations via inverse Fourier transform}
\label{subsec:real_fpe_orig}

A direct way to obtain real-valued position vectors from the standard \gls{FPE} construction is to apply the inverse Fourier transform to the complex-domain representation in \cref{eq:fpe_real_ifft}. Such real-valued vectors can then be used with Hadamard product binding, as done, for example, in time-series classification applications~\cite{Schlegel2022b}. 

A special case arises for \(\beta=0\): the complex vector is \(\mathbf{c}(t,0)=[1,\dots,1]\), and the inverse Fourier transform yields a vector concentrated in a single component, \(\mathbf{p}(t,0)=[1,0,0,\dots]\).
While this vector acts as an identity element under circular convolution, it is unsuitable for Hadamard product binding because all but one component of the position vector are zero, causing the corresponding components of the bound vector to be zero as well.

To avoid this issue, we shift the exponent in the complex-domain representation:
\[
\mathbf{c}(t,\beta)
=
\left[
1,\,
e^{i\omega_2(\beta t + 1)},\,
\dots,\,
e^{i\omega_{D/2}(\beta t + 1)},\,
1,\,
e^{-i\omega_{D/2}(\beta t + 1)},\,
\dots,\,
e^{-i\omega_2(\beta t + 1)}
\right].
\]

With this offset, \(\mathbf{c}(t,0)=\mathbf{c}(0,0)\) for all \(t\), but the resulting real-valued vector \(\mathbf{p}(t,0)\), produced by \cref{eq:fpe_real_ifft}, is no longer concentrated in a single component. This makes the representation suitable for Hadamard product binding (see also previous works such as~\cite{Schlegel2022b}.

\paragraph{Hadamard compatibility (R1):}

After applying the inverse Fourier transform by \cref{eq:fpe_real_ifft}, the resulting vectors $\mathbf{p}(t,\beta)$ lie in $\mathbb{R}^D$ and can, therefore, be used directly with real-valued Hadamard product binding.

\paragraph{Similarity kernel (R2):}

The exponent shift introduces only a constant offset, replacing \(\beta t\) by \(\beta t+1\). Since the similarity between two positions depends only on their difference, this constant offset cancels in pairwise comparisons. Consequently, the induced similarity kernel retains the same functional form as described in \cref{subsec:sim_kernel}, preserving graded similarity between nearby positions.

\paragraph{Norm stability (R3):}

The vectors produced by the inverse Fourier transform maintain 
constant norms equal to $\sqrt D$ across positions due to the unit-magnitude structure of the complex domain representation. 

\paragraph{Shift equivariance (R4):}

A limitation of this formulation is that shift equivariance is not preserved under real-valued Hadamard product binding. 
In standard \gls{FPE}, shifts are realized by multiplicative binding with the shift vector in the complex domain, or equivalently by circular convolution in the real domain. 
By contrast, applying the Hadamard product to the real-valued vectors obtained after the inverse Fourier transform does not implement the required complex-domain multiplication~\cite{Frady2021a}. Therefore, in general,
\[
\mathbf{p}(t+\delta,\beta)
\neq
\mathbf{p}(\delta,\beta)\odot\mathbf{p}(t,\beta).
\]

\paragraph{Computational efficiency (R5):}

Generating the position vectors requires exponentiation in the complex domain followed by the inverse Fourier transform. Although efficient IFFT implementations reduce the cost below quadratic complexity, they still require \(\mathcal{O}(D\log D)\) operations. This introduces additional computational overhead compared with the purely real-valued \(\mathcal{O}(D)\) formulations considered below.

Variant~A, therefore, serves primarily as a reference formulation. It preserves the standard \gls{FPE} similarity kernels and produces real-valued position vectors compatible with Hada\hyp mard product binding, but it does not provide an algebraic shift operator in the real-valued representation space.

\subsection{Variant~B: Real-valued Sinusoid positional representations:}
\label{subsec:real_FPE_sinusoid}

This variant generates real-valued position vectors directly, without  using the inverse Fourier transform, 
based on the \gls{RFF}~\cite{Rahimi2008}. 
Using the angular frequencies $\boldsymbol{\omega}=[\omega_1,\omega_2,\dots \omega_{D}]$ sampled to control the kernel shape (cf.\ \cref{fig:similarity_timesteps}) from the appropriate $\mathcal{P}$ , we define the position vector $\boldsymbol{\psi}(t,\beta)$ as:

\begin{equation}
\boldsymbol{\psi}_{\mathrm{sin}}(t,\beta)\!=\!\sqrt{\frac{1}{D}}\!
\begin{bmatrix}
\sin\!\left(\omega_1 \beta t\right)\\
\vdots \\
\sin\!\left(\omega_{D
} \beta t\right) 
\end{bmatrix}
;
\boldsymbol{\psi}_{\mathrm{cos}}(t,\beta)\!=\!\sqrt{\frac{1}{D}}\!
\begin{bmatrix}
\cos\!\left(\omega_1 \beta t\right) \\
\vdots \\
\cos\!\left(\omega_{D
} \beta t\right) \\
\end{bmatrix}
;
\boldsymbol{\psi}(t,\beta)\!=\!
\begin{bmatrix}
\boldsymbol{\psi}_{\mathrm{sin}}(t,\beta)\\
\boldsymbol{\psi}_{\mathrm{cos}}(t,\beta))
\end{bmatrix}.
\label{eq:sinusoid_fpe}
\end{equation} 

This representation corresponds to a real-valued decomposition of the complex exponential feature map. 
Though both sine and cosine components are used, we call this representation ``Sinusoid'' for brevity and to distinguish it from Variant C.
The induced similarity between encoded positions approximates the corresponding 
shift-invariant kernel (following the scalar version of
the \gls{RFF} formulation):


\begin{equation}
k(
{t_1},
{t_2}) \approx \boldsymbol{\psi}
(
{t_1},\beta)^\top \boldsymbol{\psi}
(
{t_2},\beta)=
\boldsymbol{\psi}_{\mathrm{sin}}(
{t_1},\beta)^\top \boldsymbol{\psi}_{\mathrm{sin}}(
{t_2},\beta)
+ 
\boldsymbol{\psi}_{\mathrm{cos}}(
{t_1},\beta)^\top \boldsymbol{\psi}_{\mathrm{cos}}(
{t_2},\beta). 
\end{equation}

\paragraph{Hadamard compatibility (R1):}

The position vectors $\boldsymbol{\psi}(t,\beta)$ are real-valued and directly compatible with the Hadamard product binding. 
This allows the Sinusoid representation to be integrated with MAP-based \gls{HDVSA} operations without requiring computations in the complex domain.

\paragraph{Similarity kernel (R2):}

Since the Sinusoid representation 
is derived from the RFF, the similarity kernel shape is determined by the distribution from which $\boldsymbol{\omega}$ is sampled.

\paragraph{Norm stability (R3):}

The positional vector norm remains 1 for any $t$, since 
$||\boldsymbol{\psi}(t,\beta)||^2 = 
\Sigma_{i=1}^D (\sin^2\!\left(\omega_{i
} \beta t\right)+\cos^2\!\left(\omega_{i
} \beta t\right))/D = 1.
$

This constant norm ensures stable behavior when the vectors are bound with a Hadamard product.

\paragraph{Shift equivariance (R4):}

A direct Hadamard product between two Sinusoid position vectors does not yield the shifted representation, since trigonometric functions do not satisfy a multiplicative shift identity: 
 $\sin(t+\delta) \neq\sin(t)\sin(\delta)$
and
 $\cos(t+\delta) \neq\cos(t)\cos(\delta)$. 
However, an exact shift-equivariant transformation can be constructed using the trigonometric addition identities on vector-component level:
\begin{equation}
\begin{aligned}
\sin(t + \delta) &= \sin(t)\cos(\delta) + \cos(t)\sin(\delta), \\
\cos(t + \delta) &= 
\cos(t)\cos(\delta) - \sin(t)\sin(\delta), \\
\end{aligned}
\end{equation}
which allows the vector of shifted position to be expressed as combinations of sine and cosine components as follows:

\begin{align}
\boldsymbol{\psi}_{\mathrm{sin}}(t + \delta,\beta) 
&= 
\boldsymbol{\psi}_{\mathrm{sin}}(t,\beta) \odot \boldsymbol{\psi}_{\mathrm{cos}}(\delta,\beta) 
+
\boldsymbol{\psi}_{\mathrm{cos}}(t,\beta) \odot \boldsymbol{\psi}_{\mathrm{sin}}(\delta,\beta),
\label{eq:sin_bind}
\\
\boldsymbol{\psi}_{\mathrm{cos}}(t + \delta,\beta) 
&= 
\boldsymbol{\psi}_{\mathrm{cos}}(t,\beta) \odot \boldsymbol{\psi}_{\mathrm{cos}}(\delta,\beta)
-
\boldsymbol{\psi}_{\mathrm{sin}}(t,\beta) \odot \boldsymbol{\psi}_{\mathrm{sin}}(\delta,\beta).
\label{eq:cos_bind}
\end{align}
and concatenation recovers the shifted position vector:

\begin{equation}
\boldsymbol{\psi}(t + \delta,\beta) =
\left[
\ \boldsymbol{\psi}_{\mathrm{sin}}(t + \delta,\beta) ^\top,\
\ \boldsymbol{\psi}_{\mathrm{cos}}(t + \delta,\beta) ^\top\
\right]^\top.
\label{eq:concat_sinusoid_shift}
\end{equation}

Therefore, \cref{eq:sin_bind,eq:cos_bind,eq:concat_sinusoid_shift} define a shift operator $\diamond$ acting on two Sinusoid representations as
\begin{equation}
\boldsymbol{\psi}(t,\beta)\diamond\boldsymbol{\psi}(\delta,\beta)
=
\boldsymbol{\psi}(t+\delta,\beta).
\end{equation}

Functionally, the operator $\diamond$ is implemented by the component-wise combinations given in \cref{eq:sin_bind,eq:cos_bind} followed by the concatenation in \cref{eq:concat_sinusoid_shift}.
\\

\noindent\textbf{Remark:}  
For exact shift equivariance on a position-encoded sequence of element vectors (bound by their position) as in \cref{eq:seq_enc}, the sine and cosine components in $\boldsymbol{\psi}(t,\beta)$ must correspond component-wise to the same angular frequency and be multiplied by the single corresponding component of the sequence element vector.  

This alignment can be naturally achieved by duplicating the element vectors, so that identical components interact with both the cosine and sine terms of the position vector: 
\begin{equation}
\tilde{\mathbf{f}}_t =
\begin{bmatrix}
\mathbf{f}_t \\
\mathbf{f}_t
\end{bmatrix}
\in \mathbb{R}^{2D}.
\end{equation}
The concatenated vector $\tilde{\mathbf{f}}_t$ is then bound to the position vector $\boldsymbol{\psi}(t,\beta)$ using the Hadamard product. See \cref{subsec:exp_equiv} for an experimental demonstration of this property.

Although the \textit{shift-equivariant transformation preserves the norm of the sequence representation}, this property can also be verified explicitly. 
Consider a single sine/cosine component pair of the sequence representation corresponding to one scalar component of the sequence elements $f_1,\dots,f_T$, obtained by \cref{eq:seq_enc}:
\[
r
=
\sum_{t=1}^T
f_t
\bigl(
\sin(\omega \beta t),
\cos(\omega \beta t)
\bigr).
\]
Equivalently, let
\[
A=\sum_{t=1}^T f_t \sin(\omega \beta t),
\qquad
B=\sum_{t=1}^T f_t \cos(\omega \beta t),
\]
such that
\[
r=(A,B),
\qquad
\|r\|^2 = A^2 + B^2.
\]
Now shift all positions by the same offset $\delta$. The shifted representation becomes
\[
r_\delta
=
\sum_{t=1}^T
f_t
\bigl(
\sin(\omega \beta t + \delta),
\cos(\omega \beta t + \delta)
\bigr).
\]
Using the angle addition identities yields
\[
\begin{aligned}
r_\delta
&=
\sum_{t=1}^T
f_t
\Bigl(
\sin(\omega \beta t)\cos(\delta)
+
\cos(\omega \beta t)\sin(\delta),
\\
&\qquad\qquad
\cos(\omega \beta t)\cos(\delta)
-
\sin(\omega \beta t)\sin(\delta)
\Bigr).
\end{aligned}
\]
Collecting terms gives
\[
r_\delta
=
\bigl(
A\cos(\delta)+B\sin(\delta),
\,
B\cos(\delta)-A\sin(\delta)
\bigr).
\]
The squared norm of the shifted representation is, therefore,
\[
\begin{aligned}
\|r_\delta\|^2
&=
\left(
A\cos(\delta)+B\sin(\delta)
\right)^2
+
\left(
B\cos(\delta)-A\sin(\delta)
\right)^2
\\
&=
A^2\bigl(\cos^2(\delta)+\sin^2(\delta)\bigr)
+
B^2\bigl(\cos^2(\delta)+\sin^2(\delta)\bigr)
\\
&=
A^2+B^2
=
\|r\|^2.
\end{aligned}
\]
Since the same argument applies independently to each of the \(D\) scalar components of the sequence representation, each corresponding sine/cosine pair preserves its squared norm. Therefore, the total squared norm, obtained by summing over all \(D\) pairs, is unchanged by the shift.

\paragraph{Computational efficiency (R5):}

In contrast to Variant~A, the Sinusoid \gls{FPE} representation operates entirely in real-valued position vector space and does not require Fourier transforms. 
Position vectors can be computed directly using trigonometric functions, improving computational complexity to $O(D)$, while preserving the desired similarity kernel structure of position vectors. \\
Variant~B thus provides a fully real-valued position and sequence representation that satisfies all design requirements, including an explicit exact shift-equivariance in the embedding space. However, we operate with $2D$-dimensional vectors, where $D$ is the dimensionality of sequence elements' vectors.

\subsection{Variant C: Real-valued Cosine-only positional representations:}
\label{subsec:real_FPE_cosine}

Variant~C simplifies the Sinusoid representation by using cosine-only features~\cite{Rahimi2008}. 
The Cosine positional vectors are given by:
\begin{equation}
\boldsymbol{\chi}(t,\beta)
=
\sqrt{\frac{2}{D}}\,
\cos\!\left(\boldsymbol{\omega}t\beta+\mathbf{b}\right),
\qquad
\mathbf{b}\sim\mathcal{U}(-\pi,\pi)^D .
\label{eq:cosine_rff}
\end{equation}

Here the cosine is applied component-wise. The factor \(\sqrt{2}\) compensates for using a single cosine feature instead of a sine--cosine pair, while the factor \(2/\sqrt{D}\) normalizes the random feature vector,
ensuring that the expected inner product used in the similarity kernel remains unchanged and the kernel approximation is preserved (see~\cite{Rahimi2008}).
The resulting similarity between the encoded positions, therefore, approximates the shift-invariant kernel 
$k(
{t_1},
{t_2})\approx 
\boldsymbol{\chi}(
{t_1})^\top 
\boldsymbol{\chi}(
{t_2})$.

\paragraph{Hadamard compatibility (R1):}

The position vectors $\boldsymbol{\chi}(t,\beta)$ are real-valued and can, therefore, be used directly with the Hadamard product. 
This makes representations compatible with MAP-based \gls{HDVSA} operations without requiring complex domain computations.

\paragraph{Similarity kernel (R2):}

The cosine representation preserves the kernel approximation property of \gls{RFF}. 
Consequently, the similarity between position vectors approximates the shift-invariant kernel determined by the distribution of angular frequencies $\omega_i$. 

\paragraph{Norm stability (R3):}

The squared norm of $\boldsymbol{\chi}(t,\beta)$ is
\begin{equation}
\|\boldsymbol{\chi}(t,\beta)\|^2
=
\frac{2}{D}
\sum_{j=1}^D
\cos^2(\omega_j \beta t+b_j).
\end{equation}
Using $2\cos^2(a)=1+\cos(2a)$, we obtain
\begin{equation}
\|\boldsymbol{\chi}(t,\beta)\|^2
=
1+
\frac{1}{D}
\sum_{j=1}^D
\cos(2\omega_j \beta t+2b_j).
\end{equation}
Thus, for finite $D$, the norm is generally not exactly constant across positions $t$. 
However, since $b_j\sim\mathcal{U}(-\pi,\pi)$, we have
\begin{equation}
\mathbb{E}_{b_j}
\left[
\cos(2\omega_j \beta t+2b_j)
\right]
=0,
\end{equation}
and, therefore,
\begin{equation}
\mathbb{E}_{\mathbf{b}}
\left[
\|\boldsymbol{\chi}(t,\beta)\|^2
\right]
=1.
\end{equation}
Consequently, the cosine-only random-phase representation preserves the norm in expectation. 
For finite $D$, the norm fluctuates around its expected value, and these fluctuations decrease with increasing $D$ due to averaging over the random phases.

\paragraph{Shift equivariance (R4):}
Unlike the Sinusoid representation in Variant~B, the cosine-only representation does not explicitly store the sine components required to reconstruct phase shifts using trigonometric identities. 
The $\cos(t+\delta)=\cos(t)\cos(\delta)-\sin(t)\sin(\delta)$ shows, that requires knowledge of \(\sin(t)\) which is explicitly absent here. 
Therefore, a shift-equivariant operator cannot be constructed within this representation space.

\paragraph{Computational efficiency (R5):}
The cosine representation operates entirely in a real-valued vector space and requires only cosine evaluations. Compared to Variant~B, it does not store an additional sine component for each angular frequency. This results in a simpler and computationally efficient encoding. Variant~C, therefore, provides a compact real-valued position vector that approximates the shift-invariant similarity kernel in expectation and supports efficient Hadamard products, but does not support an exact algebraic shift-equivariant operator within the stored representation space.

\paragraph{Remark -- Approximation accuracy:}

The kernel approximation error depends on the variance of the corresponding kernel estimator. 
For the same number of sampled angular frequencies \(D\), the Cosine-only representation uses \(D\) vector components, whereas the Sinusoid representation stores one sine--cosine pair per frequency, resulting in \(2D\) vector components in total.

For two positions \(t_1\) and \(t_2\), define
\[
V(t_1-t_2)
=
\operatorname{Var}_{\omega}
\left[
\cos\bigl(\omega(t_1-t_2)\bigr)
\right].
\]
The estimator variances are then given by
\[
\operatorname{Var}_{\sin\text{-}\cos}
=
\frac{V(t_1-t_2)}{D},
\qquad
\operatorname{Var}_{\cos}
=
\frac{V(t_1-t_2)+\frac{1}{2}}{D}.
\]
Hence,
\[
\frac{
\operatorname{Var}_{\cos}
}{
\operatorname{Var}_{\sin\text{-}\cos}
}
=
1+\frac{1}{2V(t_1-t_2)}.
\]

Therefore, for the same number of sampled frequencies, the Cosine-only representation exhibits a larger kernel approximation variance. 
The Sinusoid representation reduces this variance by explicitly encoding both phase components, at the cost of doubling the vector dimensionality.


\subsection{Summary}

The three variants differ in how the position vectors are constructed and how they satisfy the design requirements introduced in \cref{sec:req}. 
Variant~A follows the standard \gls{FPE} formulation and approximates the original similarity kernel, but requires Fourier transforms and does not support shift-equivariant operations under the Hadamard product. 
Variant~B provides a fully real-valued Sinusoid representation that preserves the kernel structure and enables an explicit shift-equivariant operator in real-valued vector space. 
Variant~C further simplifies the formulation by using cosine-only representations, resulting in a compact and efficient real-valued encoding that approximates the shift-invariant similarity kernel, but sacrifices exact shift equivariance.
A comparison of the variants and their properties is summarized in \cref{tab:fpe_summary}.

\begin{landscape}
\begin{table}[p]
\centering
\small
\caption[Comparison of sequence representation approaches]{Comparison of sequence representation approaches: implementation, equivariance, and sequence encoding approach.}
\label{tab:fpe_summary}
\renewcommand{\tabularxcolumn}[1]{>{\raggedright\arraybackslash}p{#1}}

\begin{tabularx}{\linewidth}{
  >{\raggedright\arraybackslash}p{2.2cm}
  L  L  L  L
}
\toprule
\textbf{Property} 
& \textbf{Circular Convolution / Complex Hadamard} 
& \textbf{Variant A: Real Hadamard + Original FPE} 
& \textbf{Variant B: Real Hadamard + Sinusoidal} 
& \textbf{Variant C: Real Hadamard + Cosine} \\
\midrule

\textbf{Representation of features $\mathbf{f}_t$ and positions $\mathbf{p}_t$ (\cref{eq:sequence_enc_bind})}
& $\mathbf{f}_t \in \mathbb{C}^D$ and $\mathbf{p}_t = \mathbf{c}(t, \beta)$, or $\mathbf{f}_t \in \mathbb{R}^D$ and $\mathbf{p}_t = \mathbf{p}(t, \beta) 
$
& $\mathbf{f}_t \in \mathbb{R}^D$ and $\mathbf{p}_t = \mathbf{p}(t, \beta)$ 
& $\mathbf{f}_t \in \mathbb{R}^D$ and $\mathbf{p}_t = \boldsymbol{\psi}(t, \beta)$
& $\mathbf{f}_t \in \mathbb{R}^D$ and $\mathbf{p}_t =  \boldsymbol{\chi}(t, \beta)$ \\ 

\hline

\textbf{Implementation of \gls{FPE}} 
& $\begin{aligned}
\mathbf{p}(t, \beta) &= \mathcal{F}^{-1}(\mathbf{c}^{\beta t}) \\
\mathbf{c}(t, \beta) &= \mathbf{c}^{\beta t},\ \text{with} \\
\mathbf{c} &= e^{i \boldsymbol{\omega}}
\end{aligned}$

& $\begin{aligned}
\mathbf{p}(t, \beta) &= \mathcal{F}^{-1}(\mathbf{c}^{\beta t + 1}) \\
\text{with} \ \mathbf{c} &= e^{i \boldsymbol{\omega}}
\end{aligned}$

& $\begin{aligned}
\boldsymbol{\psi}(t,\beta) &= 
[
\sin(\boldsymbol{\omega} t\beta),
\\&
\quad \cos(\boldsymbol{\omega} t\beta)]
\end{aligned}$

& $\begin{aligned}
\boldsymbol{\chi}(t,\beta) &= \sqrt{2}\cos(\boldsymbol{\omega} t\beta + \mathbf{b}),
\\&
b_i \sim \mathcal{U}[-\pi,\pi]
\end{aligned}$ \\

& \multicolumn{4}{c}{\raggedright
with $\omega_i \overset{\text{i.i.d.}}{\sim} \mathcal{P}$,
for $i=1,\ldots,D$
} \\

\hline
\addlinespace

\textbf{Hadamard binding
compatibility (R1)} 
& $\odot$ (complex Hadamard) or $\circledast$ (circular convolution) 
& $\odot$ (real-valued Hadamard) 
& $\odot$ (real-valued Hadamard) 
& $\odot$ (real-valued Hadamard) \\

\hline
\addlinespace

\textbf{Similarity kernel approximation (R2)} 
& preserved via complex exponential representation
& preserved (constant offset in exponent)
& preserved via RFF approximation
& preserved via RFF approximation \\

\hline
\addlinespace

\textbf{Position vector norm 
(R3)} 
& constant due to unit-magnitude complex phasors
& constant after inverse Fourier transform
& constant by Sinusoidal embedding
& approximately constant by Cosine embedding \\

\hline
\addlinespace

\textbf{Shift equivariance (R4)} 
& Satisfied: $\mathbf{p}(t + \delta) = \mathbf{p}(t) \circledast \mathbf{p}(\delta)$ \newline
$\mathbf{c}(t + \delta) = \mathbf{c}(t) \odot \mathbf{c}(\delta)$

& Not satisfied: $\mathbf{p}(t + \delta) \neq \mathbf{p}(t) \odot \mathbf{p}(\delta)$ 

& Satisfied: $\boldsymbol{\psi}(t + \delta) = \boldsymbol{\psi}(t) \diamond \boldsymbol{\psi}(\delta)$ 

& Not satisfied: $\boldsymbol{\chi}(t + \delta) \neq \boldsymbol{\chi}(t) \odot \boldsymbol{\chi}(\delta)$ \newline
$\boldsymbol{\chi}(t + \delta) \neq \boldsymbol{\chi}(t) \diamond \boldsymbol{\chi}(\delta)$ \\

\hline
\addlinespace

\textbf{Encode temporal sequence $[\mathbf{f}_1,\dots,\mathbf{f}_T]$} 

& if $\mathbf{f}_t \in \mathbb{R}^D$: \newline
$\mathbf{r} = \sum_{t=1}^{T}\mathbf{f}_t \odot \mathbf{c}(t,\beta) $ \newline
if $\mathbf{f}_t \in \mathbb{C}^D$: \newline
$\mathbf{r} = \sum_{t=1}^{T}\mathbf{f}_t \circledast \mathbf{p}(t,\beta) $

& $\mathbf{r} = \sum_{t=1}^{T}\mathbf{f}_t \odot \mathbf{p}(t,\beta) $

& $\mathbf{r} = \sum_{t=1}^{T} \mathbf{f}_t \odot \boldsymbol{\psi}(t,\beta)$

& $\mathbf{r} = \sum_{t=1}^{T}\mathbf{f}_t \odot \boldsymbol{\chi}(t,\beta)$ \\

\hline

\textbf{Shift sequence by $\delta$} 
& $\mathbf{r}_\delta = \mathbf{p}(\delta,\beta)\circledast \mathbf{r}$ \newline
$\mathbf{r}_\delta = \mathbf{c}(\delta,\beta)\odot \mathbf{r}$

& Not supported

& $\mathbf{r}_\delta = \boldsymbol{\psi}(\delta,\beta)\diamond \mathbf{r}$

& Not supported \\

\hline

\textbf{Computational efficiency (R5)} 
& circular convolution via FFT: $\mathcal{O}(D\log D)$ 
& requires exponentiation and inverse FFT: $\mathcal{O}(D\log D)$
& purely real operations: $\mathcal{O}(D)$
& purely real operations: $\mathcal{O}(D)$ \\

\bottomrule
\end{tabularx}
\end{table}
\end{landscape}

\section{Experimental evaluation}
\label{sec:exp}

This section presents an experimental evaluation of the proposed real-valued positional vector variants introduced in \cref{sec:methods}, which enable the use of efficient real-valued Hadamard product binding for sequence representation. 
The experiments aim to assess both the empirical performance of the investigated variants and their properties with respect to the design requirements specified in \cref{sec:req}. 
In particular, we analyze their effectiveness in time-series classification tasks and investigate the shift equivariance property (R4) enabled by the Sinusoid
representation (Variant~B).

\subsection{Time-series classification experiment}
\label{subsec:exp_tsc}

\begin{table}[tb]
\small
\caption{Datasets with more than $\pm 5\%$ relative change in ACC
between ROCKET variants~\cite{Dempster2021,Dempster2023} and the corresponding HDC-based approaches. 
``\texttt{Num DS}'' denotes the number of datasets in the respective subset. 
All values, except for AUPRC, are reported as percentages. 
The last four columns show the relative changes in ACC and Error. 
Each HDC model refers to the HDC-based implementation using a particular positional vector variant applied to the corresponding original (non-HDC) model in the same row.}
\begin{tabularx}{\linewidth}{p{2.9cm} p{0.4cm} p{0.5cm}  *{2}{X}  *{4}{X}}
\toprule
\textbf{Model} & \textbf{HDC} & \textbf{Num DS} & \textbf{Mean ACC} & \textbf{Worst ACC} 
& \multicolumn{2}{c}{\parbox{2cm}{\centering \textbf{Relative Change in ACC}}}
& \multicolumn{2}{c}{\parbox{2cm}{\centering \textbf{Relative Change in Error}}} \\
\cmidrule(lr){6-7} \cmidrule(lr){8-9}
& & & & 
& \textbf{Mean} & \textbf{Max} 
& \textbf{Mean} & \textbf{Min} \\
\midrule
MiniROCKET &  & 38 & 80.79 &  29.27 &  \\
HDC original HRR & HRR & 38 & 82.22 &  32.61 & 1.43 & 14.42 & -8.0 & -40.0 \\
Standard FPE & MAP & 38 & 82.23 & 30.56 & 1.44 & 14.00 & -6.0 & -77.0 \\
Sinusoid FPE & MAP & 38 & 82.22 &  30.36 & 1.43 & 14.06 & -6.0 & -76.0 \\
Cosine FPE & MAP & 38 & \textbf{82.24} & 30.55 & 1.45 & 14.77 & -6.0 & -77.0 \\
\hline
MultiROCKET-HYDRA &  & 38 & 82.72 &  36.73  \\
Standard FPE & MAP & 38 & 83.80 &  36.28 & 1.08 & 23.58 & -6.0 & -62.0 \\
Sinusoid FPE & MAP & 38 & \textbf{83.86} & 35.68 & 1.14 & 23.58 & -7.0 & -62.0 \\
Cosine FPE & MAP & 38 & 83.81 & 35.62 & 1.09 & 23.58 & -6.0 & -62.0 \\

\bottomrule
\end{tabularx}

\label{tab:summary_results_diff_FPE}
\end{table}

To evaluate the empirical performance of the proposed real-valued variants, we conduct a time-series classification experiment using the HDC-MiniROCKET and HDC-\hspace{0pt}Multi\hspace{0pt}ROCKET-\hspace{0pt}HYDRA framework similar to~\cite{Schlegel2022b,Schlegel2025a}. 
In this framework, state-of-the-art Mini\-ROCKET~\cite{Dempster2021} or Multi\hspace{0pt}ROCKET/\hspace{0pt}HYDRA~\cite{Dempster2023,Tan2022} sequence element vectors are bound with temporal position vectors using a real-valued Hadamard product as defined in \cref{eq:method_seq_enc}. 
This setup allows a direct comparison of the position encodings introduced in \cref{sec:methods} under identical sequence encoding conditions.

We compare the following position encoding approaches:

\begin{itemize}
\item \textbf{Standard FPE (Variant~A)}: real-valued position vectors obtained via inverse Fourier transform as described in \cref{subsec:real_fpe_orig},
\item \textbf{Sinusoid (Variant~B)}: the real-valued Sinusoid representation derived from RFF as introduced in \cref{subsec:real_FPE_sinusoid},
\item \textbf{Cosine (Variant~C)}: the Cosine representation presented in \cref{subsec:real_FPE_cosine}.
\end{itemize}

For all variants, we sampled the angular-frequency components independently as $\omega_i \sim \mathcal{U}(-\pi,\pi)$, yielding the \texttt{sinc} similarity kernel.
The experiments use MiniROCKET and Multi-ROCKET-HYDRA models evaluated on datasets from the UCR time-series archive~\cite{Dau2019}. 
Following the procedure in~\cite{Schlegel2025a}, the kernel width parameter $\beta$ was selected by grid search via cross-validation to determine the best-performing value for each configuration. 
The search was performed over a logarithmically spaced range defined as $\beta = 2^i - 1$ with $i \in \{0, 0.5, 1, 1.5, 2, 2.5, 3\}$, resulting in $\beta \in \{0, 0.41, 1.00, \allowbreak 1.83, 3.00, 4.66, 7.00\}$. 
All results are averaged over 30 repetitions with different random seeds 
to account for variability in the training procedure. 
Accuracy (ACC) and Error (ERR) are used as the primary evaluation metrics. 
When comparing with baseline models, we measure relative changes in accuracy and error following~\cite{Schlegel2025a}:
\begin{equation}
\begin{aligned}
\texttt{relative ACC change} &= \frac{\mathrm{ACC}_{\mathrm{HDC}}}{\mathrm{ACC}_{\mathrm{orig}}} - 1, \\
\texttt{relative Error change} &= \frac{1 - \mathrm{ACC}_{\mathrm{HDC}}}{1 - \mathrm{ACC}_{\mathrm{orig}}} - 1.
\end{aligned}
\label{eq:relative_changes}
\end{equation}

\Cref{tab:summary_results_diff_FPE} summarizes the classification results. 
For completeness, we additionally report the results for the HRR~\cite{Plate1993a} model combined with the standard \gls{FPE} (Variant A) for the MiniROCKET model (using circular convolution for binding with the Fourier transform). 
For the remaining experiments, we focus on the MAP-based implementations discussed in the previous section.
To highlight the differences between the \gls{FPE} variants, the table lists only datasets where the relative change in ACC or Error between the original ROCKET models and the corresponding HDC-based models exceeds $\pm 5\%$ (see also~\cite{Schlegel2025a} for the used metrics). 
Across these datasets, the overall performance of the variants is comparable. 
For MiniROCKET, the HDC-based models improve the mean accuracy by approximately $1.4\%$ on average, with maximum improvements of up to $14.8\%$. 
The mean relative error reduction is up to $-8\%$, reaching up to $-77\%$ for one dataset.

For MultiROCKET-HYDRA, the mean relative accuracy improvement is about $1.1\%$, with maximum gains of up to $23.6\%$ and relative error reductions of up to $62\%$. 
Among the evaluated position encodings, the Sinusoid representation (Variant~B) achieves the highest mean accuracy for MultiROCKET-HYDRA, while the Cosine representation (Variant~C) and the inverse-Fourier implementation (Variant~A) perform similarly. \\

These results indicate that the proposed real-valued variants maintain the representational quality of the standard \gls{FPE} representations while enabling efficient sequence encoding compatible with Hadamard product binding. 
In particular, Variant~B combines competitive classification performance with the additional shift equivariance property described in \cref{subsec:real_FPE_sinusoid}. The Sinusoid representation was also used as position encoding in~\cite{Schlegel2025a}.

\subsection{Distribution of position vector components}
\label{subsec:exp_dist}

\begin{figure}[t]
  \centering
  \includegraphics[width=0.9\linewidth]{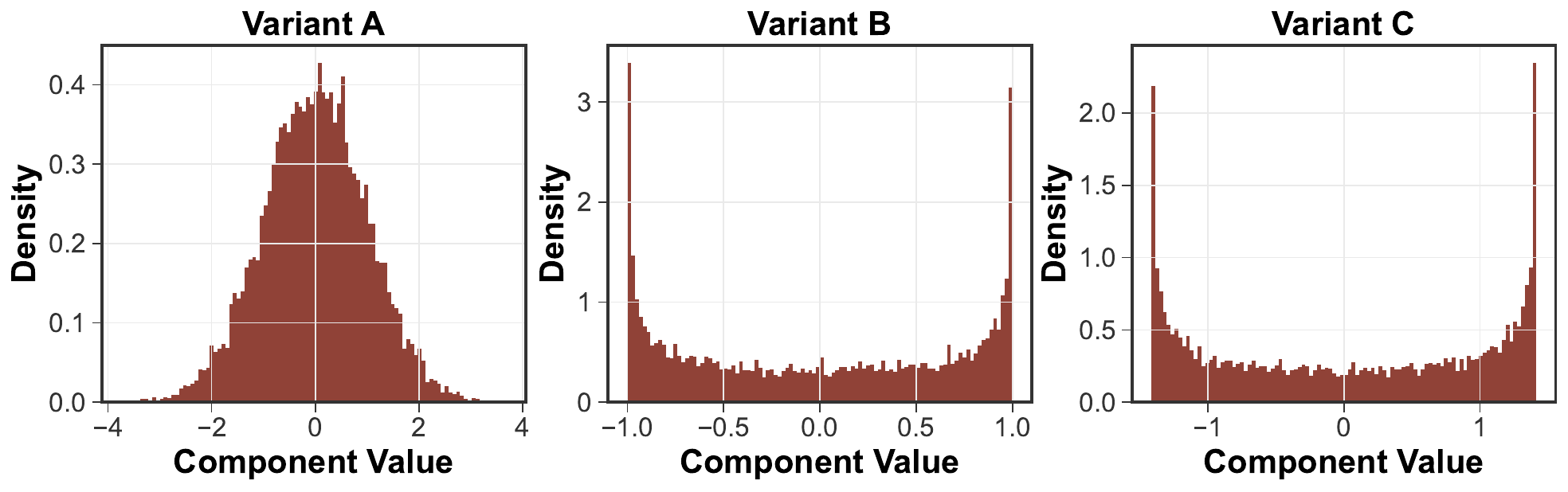}
\caption{Histograms of component values for different \gls{FPE} position encodings for the \texttt{sinc} similarity kernel. 
Each histogram is computed over all vector components for one scalar input encoded with \gls{FPE}. 
Left: Standard \gls{FPE} (inverse Fourier transform, Variant A), middle: Sinusoid (Variant~B), right: Cosine (Variant C).}
  \label{fig:dist_fpe_variants}
\end{figure}

Another aspect influencing the behavior of binding with the Hadamard product is the component value distribution of the generated position vectors.
\Cref{fig:dist_fpe_variants} shows these distributions as histograms for the three variants.
The histograms are computed for the \texttt{sinc} kernel (\cref{fig:similarity_timesteps})
with a $\beta=1$ and positions between $-1$ and $1$. 
All vector components of all generated position vectors are aggregated into the histogram.
The standard \gls{FPE} (Variant~A) produces approximately Gaussian-distributed component values, since each real-valued component produced by the inverse Fourier transform sums $\approx D/2$ random cosine contributions arising from conjugate Fourier-frequency pairs.
In contrast, the Sinusoid (Variant~B) and Cosine-only (Variant~C) representations produce bounded trigonometric values. 
When the phases are approximately uniformly distributed modulo \(2\pi\), these values follow an arcsine-type distribution, with higher density near the limiting values and lower density near \(0\). For Variant~B, the sine and cosine components are bounded in \([-1,1]\), whereas for Variant~C, the \(\sqrt{2}\)-scaled cosine components are bounded in \([-\sqrt{2},\sqrt{2}]\).

This difference is relevant for the Hadamard product, which performs component-wise multiplication between element vectors and position vectors. Position-vector components close to zero suppress the corresponding bound components, whereas components close to \(-1\) or \(1\) approximately preserve their magnitude. One might expect this 
to influence the quality of the resulting sequence representation and, consequently, classification accuracy. However, in our experiments we did not observe a pronounced effect on the results, suggesting that the practical impact of these distributional differences is limited in the tested settings and deserves further investigation in future work.

\subsection{Shift equivariance evaluation}
\label{subsec:exp_equiv}

This experiment demonstrates the shift equivariance property enabled by the Sinusoid representation (Variant~B). 
As described in \cref{subsec:real_FPE_sinusoid}, this representation admits a defined shift operator $\diamond$ that allows temporal shifts to be applied directly in the embedding space without recomputing the full sequence encoding.

\begin{figure}[tb]
    \centering
    \includegraphics[width=0.9\textwidth]{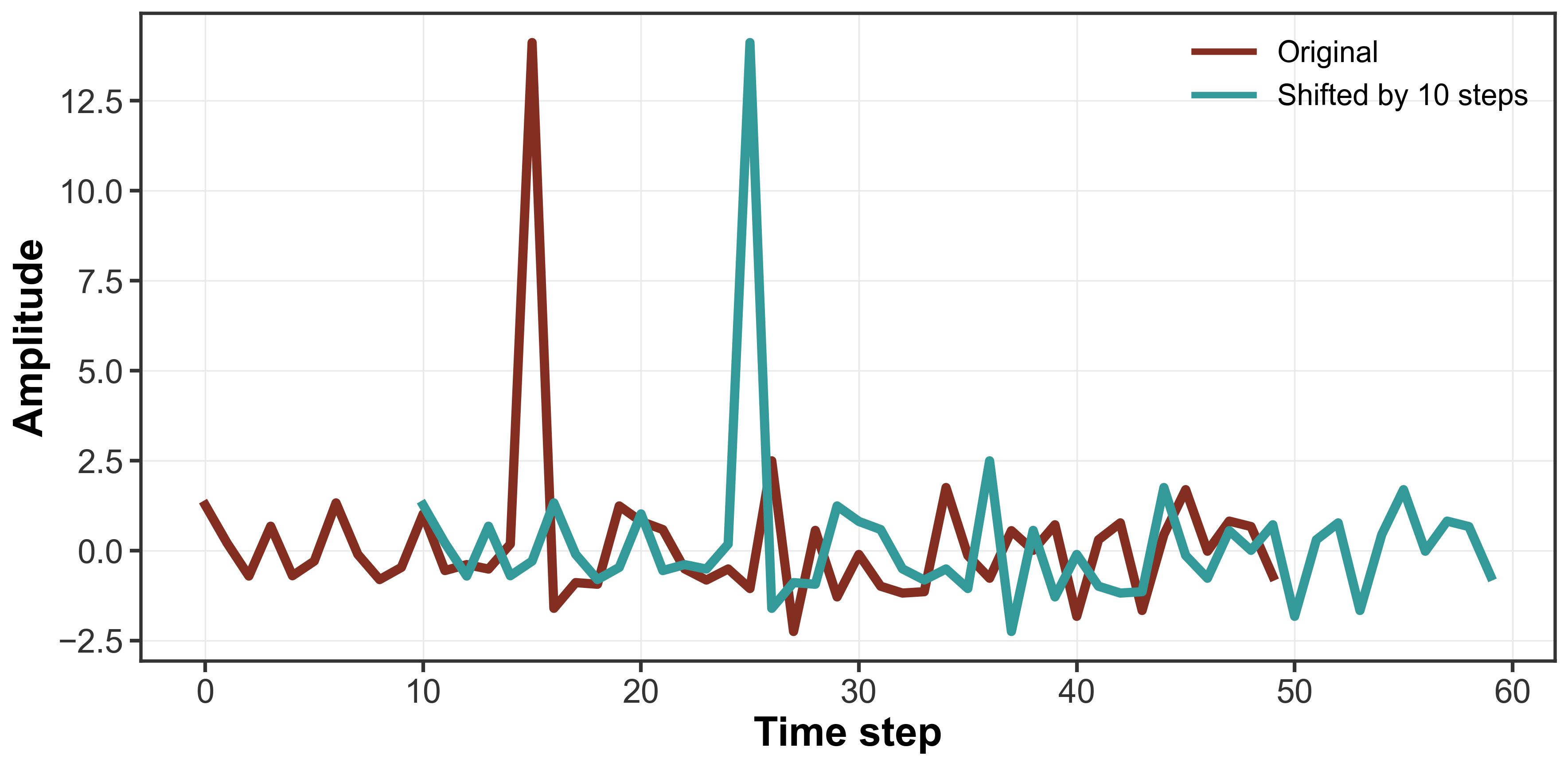}
    \caption[Original and shifted synthetic signal]
    {Example evaluation of the shift equivariance property of the Sinusoid representation in the HDC-MiniROCKET framework. A synthetic signal with a single temporal peak and the same signal shifted by 10 time steps.}
    \label{fig:shifted_signal}
\end{figure}

To illustrate this behavior, we consider a simple synthetic signal consisting of a single peak at a specific temporal position. 
The signal and a version shifted by 10 time steps are shown in \cref{fig:shifted_signal}.

Two approaches are compared:

\begin{enumerate}
\item \textbf{Re-encoded shift}: the sequence representation is recomputed after shifting the signal, producing
\begin{equation}
\mathbf{r}_{\mathrm{re},s} =
\sum_{t=1}^{T} \mathbf{p}_{t+s} \odot \mathbf{f}_t .
\end{equation}

\item \textbf{Equivariant shift}: the shift operator is applied directly to the sequence representation
\begin{equation}
\mathbf{r}_{\diamond,s} =
\mathbf{r} \diamond \mathbf{p}_s ,
\end{equation}
where $\mathbf{r} = \sum_{t=1}^{T} \mathbf{p}_t \odot \mathbf{f}_t$.
\end{enumerate}

\begin{figure}[htb]
    \centering
    \includegraphics[width=0.9\textwidth]{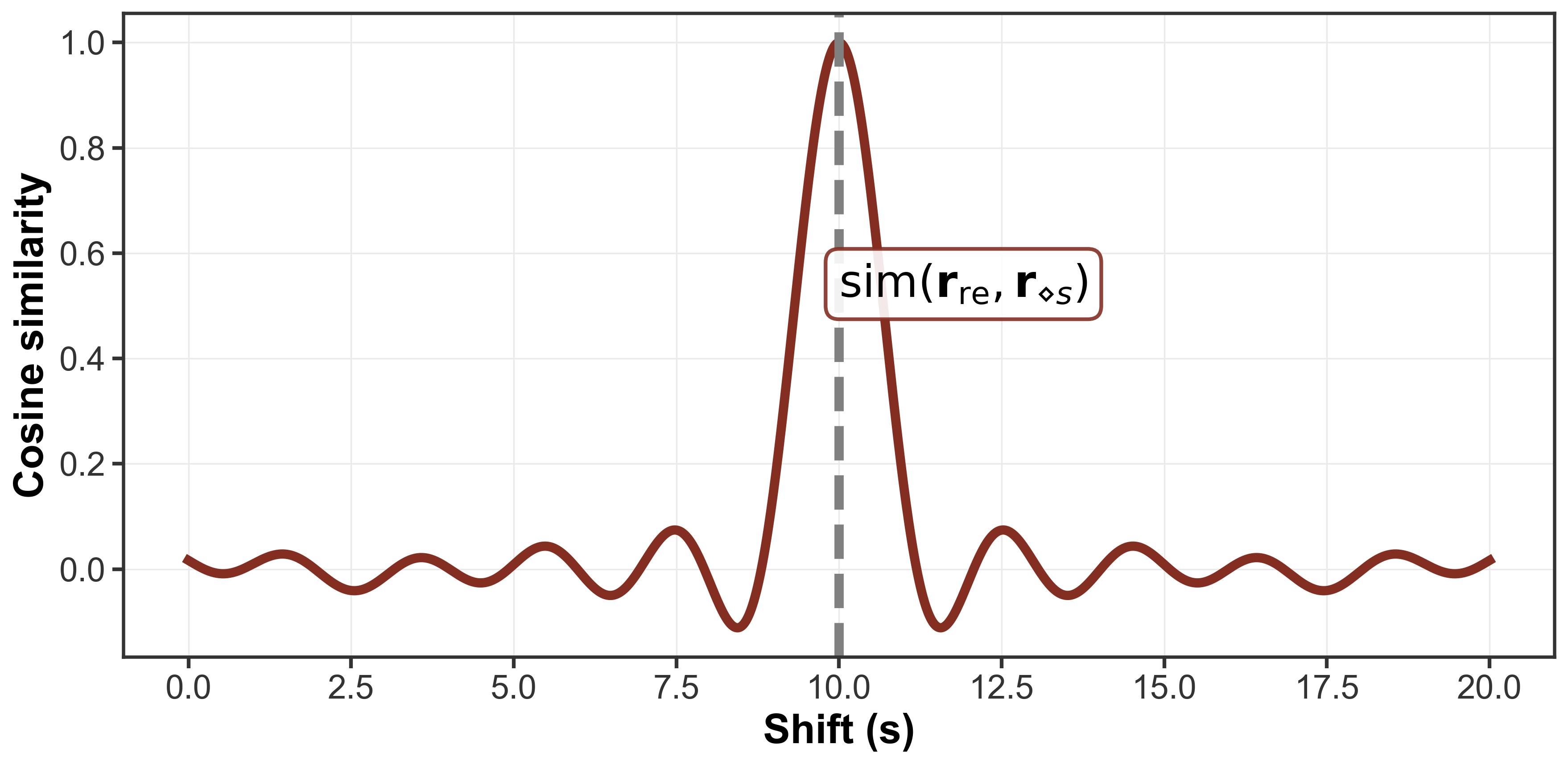}
    \caption[Demonstration of shift equivariance in HDC-MiniROCKET]{Cosine similarity between the explicitly re-encoded representation $\mathbf{r}_{\mathrm{re},s}$ and the equivariantly shifted representation $\mathbf{r}_{\diamond,s}$ for different candidate shift values \(s\).
    The peak value of \(1.0\) at the true shift \(s=10\) confirms that the equivariant transformation produces the same representation as explicit re-encoding.
    }
    \label{fig:shift_equivariance_demo}
\end{figure}

To verify equivariance, we compared both resulting sequence representations and verified their identity. 
Also, we evaluated
the cosine similarity between the re-encoded representation $\mathbf{r}_{\mathrm{re},s}$ and the equivariantly shifted representation $\mathbf{r}_{\diamond,s}$ for a continuous range of shift values. 
The results in \cref{fig:shift_equivariance_demo}
effectively demonstrate the similarity kernel of the employed position encoding, with the 
similarity of 1.0 at the correct shift magnitude ($s=10$).

These results confirm that, as expected, the Sinusoid (Variant~B) representation preserves the shift equivariance property in practice. 
Consequently, temporal shifts can be applied efficiently in the embedding space without recomputing the full sequence representation. \\

\paragraph{Computational effort:}

The proposed shift operator \(\diamond\) performs shift operations directly in real-valued space with linear complexity \(\mathcal{O}(D)\). In contrast, the standard \gls{FPE} relies on circular convolution, which is typically implemented efficiently using FFT--IFFT operations and requires \(\mathcal{O}(D\log D)\) time. By avoiding Fourier transforms, the proposed variant removes the logarithmic factor in the computational complexity of shift operations. Experimentally, we observed an approximately \(5\times\) speed-up for shift operations at \(D=1000\).

\section{Discussion}
\label{sec:discussion}

The proposed real-valued similarity-preserving position encoding variants are compatible with efficient Hadamard product binding in the Multiply-Add-Permute model. 
The results highlight several trade-offs between properties of representations and computational efficiency.

Variant~A (real-valued \gls{FPE} with Hadamard product) follows the standard \gls{FPE} formulation and, therefore, preserves the corresponding similarity kernel and representation behavior. 
The position vectors are generated via exponentiation in the complex domain followed by the inverse Fourier transform, \cref{eq:conv_power}. 
In practice, this computation only needs to be performed once when constructing the position vectors and, therefore, it does not significantly affect the cost of encoding. 
However, this variant does not preserve shift equivariance under Hadamard product binding, and therefore, temporal shifts require recomputing the sequence representation. 
Furthermore, for the angular-frequency distribution (uniform in $[-\pi,\pi]$) used in our experiments, the resulting position vectors exhibit approximately Gaussian-distributed components centered at zero (cf. \Cref{fig:dist_fpe_variants}, left), so that many position-vector components have small magnitudes. 
Under Hadamard product binding, these small-magnitude components attenuate the corresponding components of the sequence-element vector during component-wise multiplication.
The practical impact of this effect deserves further investigation in future work.

Variant~B introduces Sinusoid position vectors derived from Random Fourier Features. This formulation preserves the similarity kernel while enabling an explicit shift operator in the real-valued vector space.
As confirmed in the experiments (cf. \Cref{fig:shift_equivariance_demo}), the encoded sequence representation can be shifted algebraically without recomputing the sequence representation. Among the proposed variants, this formulation, therefore, fully satisfies the design requirements introduced in \cref{sec:req}, combining binding with the Hadamard product with shift-equivariant behavior.
Additionally, the Sinusoid \gls{FPE} representation yields an arcsine-type distribution of the position vector component values, with density increasing toward the limiting values \(-1\) and \(1\) (cf. \Cref{fig:dist_fpe_variants}, center).
This reduces the prevalence of near-zero components compared with the zero-centered Gaussian-like distribution observed for Variant~A.

Lastly, Variant~C further simplifies the representation by using cosine-only components. 
This formulation retains the ability to approximate the shift-invariant similarity kernel while providing a particularly simple and efficient implementation in a real-valued vector space.
However, since the sine components are removed, the 
representation does not support a shift-equivariant operator for algebraically shifting the encoded sequence representation. This variant may, therefore, be preferable in applications where computational simplicity is more important than equivariance.

Like Variant~B, the resulting vectors have relatively few values near zero (cf. \Cref{fig:dist_fpe_variants}, right).
This makes them well-suited for binding with the Hadamard product, since fewer position-vector components strongly attenuate the corresponding sequence-element components during component-wise multiplication.

\paragraph{Connection to Rotary Positional Encoding:}

The Sinusoid \gls{FPE} formulation in Variant~B is conceptually related to Rotary Positional Encoding (RoPE), which is used in transformer neural networks~\cite{Su2024}. 
RoPE represents positional information through rotations in pairs of vector components, so that the inner product between position-transformed vectors depends on their relative positional difference.
A similar trigonometric phase mechanism appears in the Sinusoid variant. 
Positions are encoded using paired sine and cosine components corresponding to frequency-dependent phase angles. 
Temporal shifts, therefore, correspond to phase shifts, allowing the shift operator $\diamond$ to transform encoded sequence representations directly in the embedding space.

Despite this mathematical similarity, the two approaches serve different purposes. 
RoPE injects positional information into the attention mechanism by rotating query and key vectors before their similarity is computed. 
In contrast, the Sinusoid variant constructs explicit position vectors that are used for binding-based sequence representations in \gls{HDVSA}. 
Consequently, RoPE operates within attention-based neural networks, whereas our approach supports compositional vector operations through binding.
This connection highlights that trigonometric phase representations provide a common mechanism for modeling relative positions across different sequence representation frameworks.
 
\paragraph{Limitations and future work:}

While the proposed variants of real-valued positional encodings enable efficient sequence encoding with Hadamard product binding, several limitations remain.
First, the Sinusoid variant requires paired cosine and sine components to preserve the exact shift equivariance. 
This increases the dimensionality of the positional representation and requires each sequence-element component to be bound to both phase components.

Second, the experiments focus primarily on time-series classification tasks using the HDC-MiniROCKET framework. 
Although these experiments demonstrate the practical feasibility of the proposed variants, further evaluation on additional sequence processing tasks would provide a broader validation of the approach. 
In particular, shift-equivariant encodings enable operations such as efficient search over temporal shifts or alignment between sequences, as explored in earlier work on \gls{HDVSA} sequence processing~\cite{Rachkovskij2022b,Rachkovskij2024}. 
Investigating such operations in combination with the proposed real-valued variants may provide further insights into their advantages for structured sequence analysis.

Finally, the shift-equivariance property is particularly relevant in streaming or continuously processed data scenarios. 
In such settings, new data arrive sequentially, and the temporal representation must be updated by shifting the existing sequence representation, similar to recurrent or state-space models. 
An equivariant encoding allows this shift to be applied directly in the representation space without recomputing the full sequence encoding, potentially enabling more efficient processing of long or continuously evolving data streams.

\section{Conclusion}
\label{sec:conclusion}

This paper investigated similarity-preserving position encodings for sequence representation in \gls{HDVSA}, motivated by Fractional Power Encoding (\gls{FPE}) and Random Fourier Features (\gls{RFF}). Standard \gls{FPE} relies on circular convolution and complex-domain computations, which are not directly compatible with the real-valued Hadamard product binding used in the Multiply-Add-Permute \gls{HDVSA} model. To address this limitation, we introduced three real-valued position-encoding variants that support efficient sequence encoding with Hadamard product binding. These variants 
approximate the shift-invariant similarity structure associated with \gls{RFF}- and \gls{FPE}-based representations while reducing the computational complexity of position binding from \(\mathcal{O}(D\log D)\) to \(\mathcal{O}(D)\).

Among the considered variants, the Sinusoid variant provides an explicit shift operator in real-valued vector space, enabling algebraic shift-equivariant transformations of encoded sequence representations. The Cosine-only variant provides a simpler alternative when computational simplicity is more important than shift equivariance. Empirical evaluation on a large collection of time-series classification tasks shows that the Sinusoid variant achieves performance comparable to, and in some cases better than, the standard formulation, while enabling efficient \(\mathcal{O}(D)\) real-valued implementation.

Overall, the results show that similarity-preserving position encodings can be integrated with Hadamard product binding in \gls{HDVSA}. In particular, the Sinusoid variant preserves the key shift-equivariance property while avoiding Fourier-domain computations, making it a promising alternative for efficient real-valued sequence representation.

\paragraph{Author contributions}
All authors participated in discussions that shaped the ideas, experimental design, and research questions of this study. KS conceived the main idea, developed the experimental software, performed the experiments, wrote the initial manuscript, and prepared the figures and tables. DAR and DK contributed to the development and refinement of the proposed approach, the experimental design, and the mathematical formulation, and substantially improved the manuscript text and figures. AL, SS, and EO contributed to the conceptual formulation of the study, provided feedback on the proposed approach and interpretation of the results, and helped improve the manuscript. The manuscript was written with input from all authors. All authors reviewed and approved the final manuscript.

\paragraph{Acknowledgements}
The authors would like to express their sincere gratitude to the researchers and contributors who created, collected, and curated the UCR Time Series Classification Archive~\cite{Dau2019}.
\newline
The work of AL and DK was supported by Knut and Alice Wallenberg Foundation under the Wallenberg Scholars program (Grant No. KAW2023.0327). DK acknowledges funding from the Swedish Strategic Research Foundation under the Future Research Leaders program (Grant No. FFL24-0111) and the Swedish Research Council under the Starting Grant program (Grant No. 2025-05421).
The work of EO and DR was supported in part by Swedish Research Council (VR) under Grant 2022-04657; in part by the National Academic Infrastructure for Supercomputing in Sweden (NAISS), funded by Swedish Research Council under Grant 2022-06725; and in part by the Intel Neuromorphic Research Community Project: Unsupervised Learning in NLP Tasks on Using Vector-Symbolic Representations on Phasor-Based Associative Memory. 
EO was also supported by Swedish Foundation for Strategic Research (SSF) under Grant SM 24-0003.

\bibliographystyle{unsrt}
\bibliography{library}


\end{document}